%% file: acl_latex.tex
\pdfoutput=1

\documentclass[11pt]{article}

\usepackage[]{acl}

\usepackage{times}
\usepackage{latexsym}

\usepackage{multirow}
\usepackage{amsmath}
\usepackage{amssymb}
\usepackage{graphicx}
\usepackage{makecell}
\usepackage{booktabs}

\usepackage{subcaption}

\usepackage{microtype}

\usepackage[T1]{fontenc}

\usepackage[utf8]{inputenc}

\usepackage{microtype}
\usepackage{xspace}
\newcommand{\oday}{\textit{1-day}\xspace}
\newcommand{\tday}{\textit{3-day}\xspace}
\newcommand{\sday}{\textit{7-day}\xspace}

\newcommand{\tchunk}{\textit{3-chunk}\xspace}
\newcommand{\schunk}{\textit{7-chunk}\xspace}
\newcommand{\tenchunk}{\textit{10-chunk}\xspace}

\title{Improving Text-based Early Prediction by Distillation from \\ Privileged Time-Series Text}

\author{Jinghui Liu$^{a,b}$ \quad
    Daniel Capurro$^a$ \quad
    Anthony Nguyen$^b$ \quad
    Karin Verspoor$^{c,a}$ \\
  $^a$ School of Computing and Information Systems, The University of Melbourne \\
  $^b$ Australian e-Health Research Centre, CSIRO \\
  $^c$ School of Computing Technologies, RMIT \\
  \texttt{jinghui.liu@student.unimelb.edu.au}\\
  \texttt{dcapurro@unimelb.edu.au,} \quad
\texttt{anthony.nguyen@csiro.au} \\
\texttt{karin.verspoor@rmit.edu.au} \\
 }

\begin{document}
\maketitle
\begin{abstract}
Modeling text-based time-series to make prediction about a future event or outcome is an important task with a wide range of applications. The standard approach is to train and test the model using the same input window, but this approach neglects the data collected  in longer input windows between the prediction time and the final outcome, which are often available during training. In this study, we propose to treat this neglected text as privileged information available during training to enhance early prediction modeling through knowledge distillation, presented as \textbf{L}earning \textbf{u}sing \textbf{P}rivileged t\textbf{I}me-s\textbf{E}ries \textbf{T}ext (LuPIET). We evaluate the method on clinical and social media text, with four clinical prediction tasks based on clinical notes and two mental health prediction tasks based on social media posts. Our results show LuPIET is effective in enhancing text-based early predictions, though one may need to consider choosing the appropriate text representation and windows for privileged text to achieve optimal performance. Compared to two other methods using transfer learning and mixed training, LuPIET offers more stable improvements over the baseline, standard training. As far as we are concerned, this is the first study to examine learning using privileged information for time-series in the NLP context.

\end{abstract}

\section{Introduction}
\input{sections/introduction}

\section{Related Work}
\input{sections/related_work}

\section{Methods}
\input{sections/method}

\section{Experiments}
\input{sections/experiment}

\section{Results}

\input{sections/results}

\section{Discussion}
\input{sections/discussion}

\section{Conclusion}
\input{sections/conclusion}

\section*{Acknowledgements}
We would like to thank the anonymous reviewers
for their helpful reviews and comments. This research was undertaken using the LIEF HPC-GPGPU Facility hosted at the University of Melbourne. This Facility was established with the assistance of LIEF Grant LE170100200.

\bibliography{custom}
\bibliographystyle{acl_natbib}



\end{document}

%% file: sections/introduction.tex
 Time-series forecasting, or early prediction, is an important machine learning task with a wide range of applications, such as weather prediction~\citep{Krasnopolsky2006-vs,Espeholt2022-sd} and stock forecasting~\citep{Xu2018-zy,Sharma2017-ss}. Predicting future events or outcomes would enable timely responses that can bring significant social and economic benefits. Meanwhile, most existing works on forecasting or early prediction use structured measurements or features as input~\citep{Steyerberg2009-jw}, and studies to leverage unstructured text to explore temporal patterns are still scarce~\citep{Assale2019-fs}. Moreover,  user-generated, domain-specific textual data, such as clinical and social media texts, can be noisy and complex to model~\citep{Baldwin2013-gb,Huang2020-td}. This creates challenges in utilizing text for early prediction.
 
\begin{figure}[t!]
     \centering 
     
     \begin{subfigure}[b]{0.23\textwidth}
         \centering
         \includegraphics[width=\textwidth]{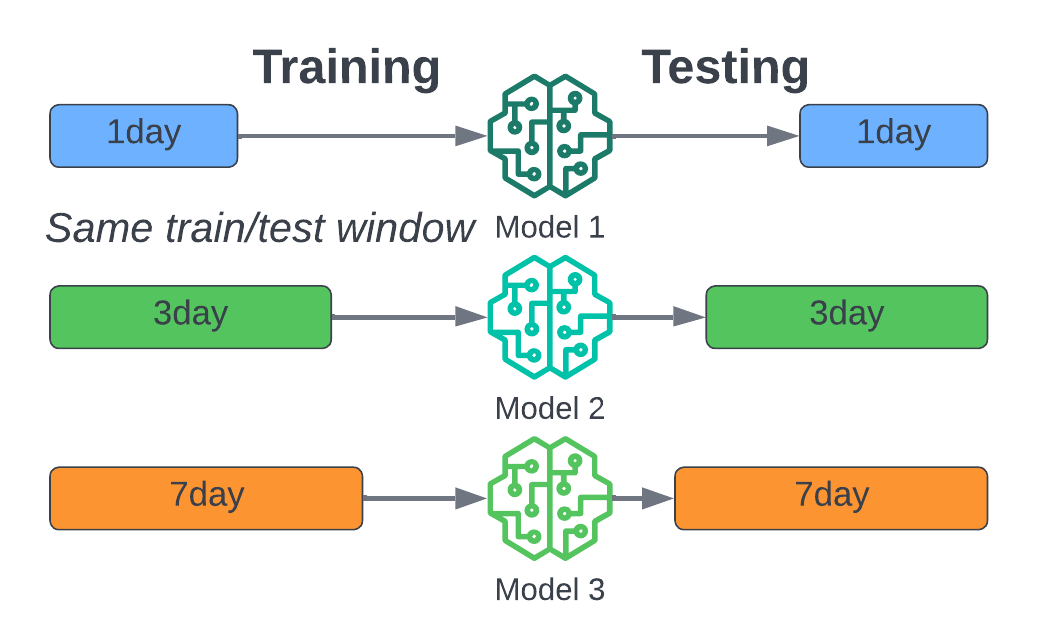}
         \caption{Standard training}
         \label{fig:base}
     \end{subfigure}
     \hfill
     \begin{subfigure}[b]{0.23\textwidth}
         \centering
         \includegraphics[width=\textwidth]{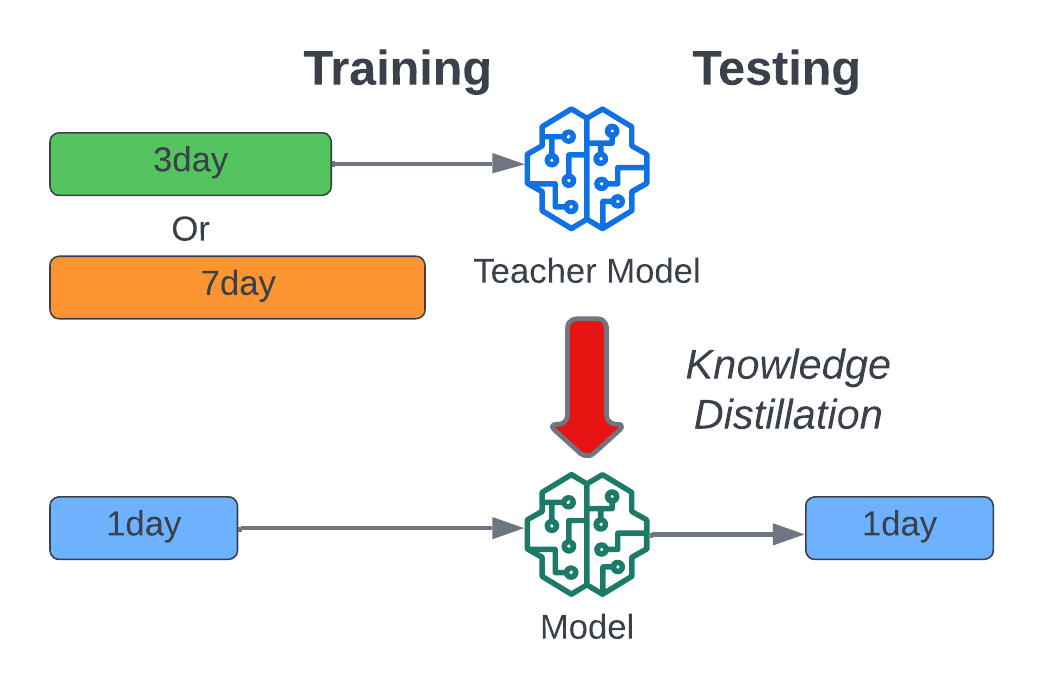}
         \caption{LuPIET}
         \label{fig:lupiet}
     \end{subfigure}
     \hfill
     \vfill
     \begin{subfigure}[b]{0.23\textwidth}
         \centering
         \includegraphics[width=\textwidth]{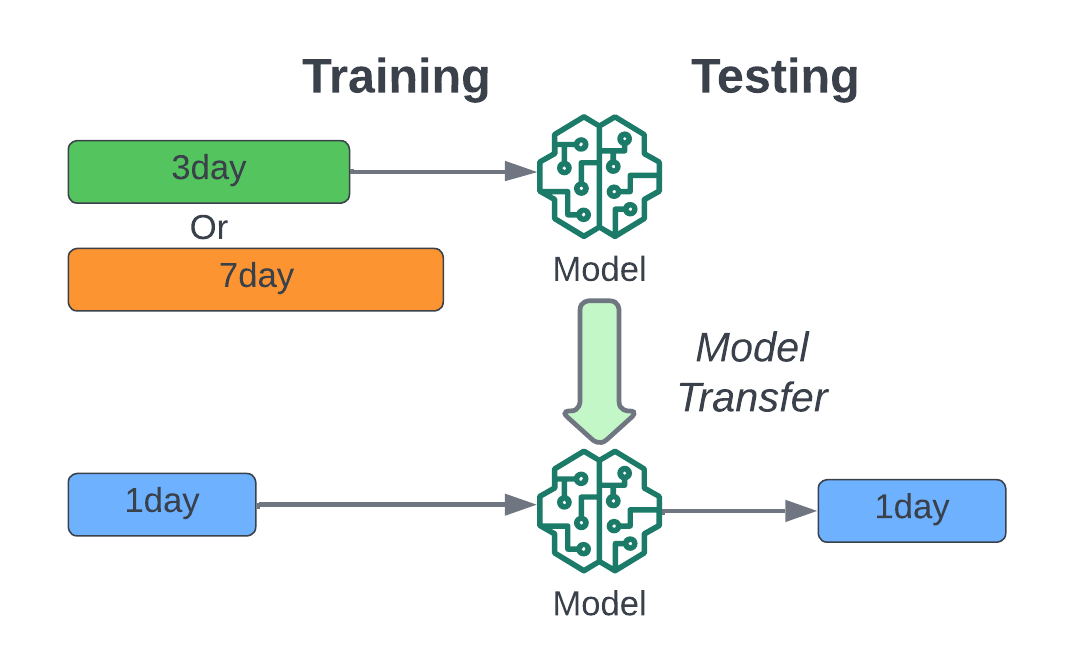}
         \caption{Transfer learning}
         \label{fig:transfer}
     \end{subfigure}
     \hfill
     \begin{subfigure}[b]{0.23\textwidth}
         \centering
         \includegraphics[width=\textwidth]{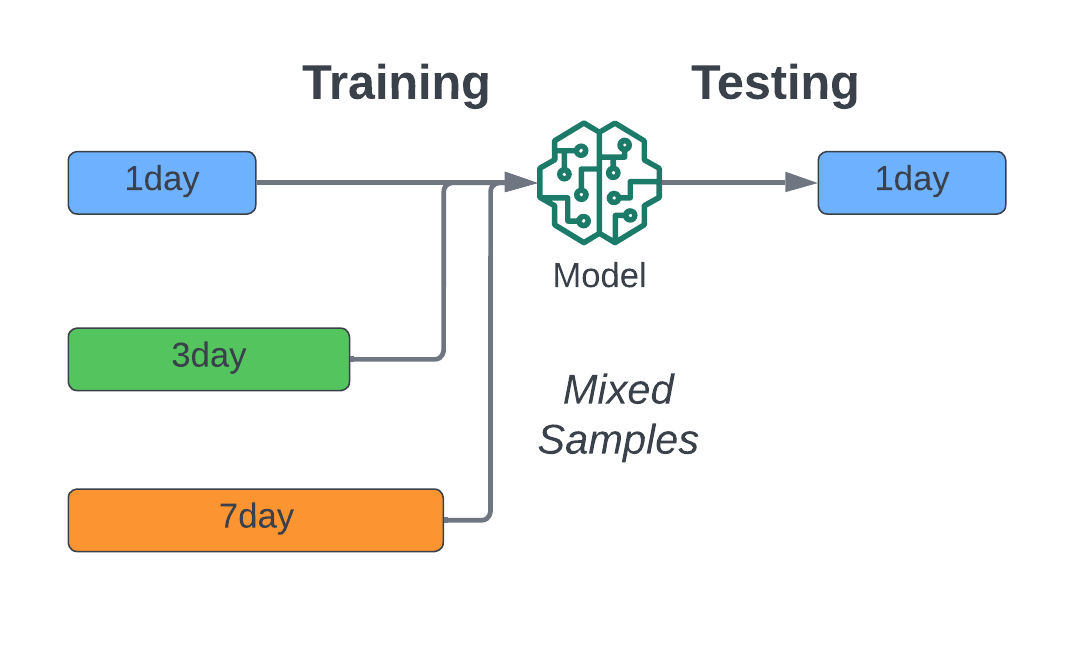}
         \caption{Mixed training}
         \label{fig:mixed}
     \end{subfigure}
     
        \caption{Different methods to leverage later data from the time-series to assist early prediction. LuPIET refers to learning using privileged time-series text in training via knowledge distillation. Here \oday is the baseline prediction window at test time. Models may leverage data from the prolonged training windows, e.g., \tday, to enhance the performance for the shorter test window.}
        \label{fig:two}
\end{figure}

The standard framework for early prediction trains and tests machine learning models using the same input window, depicted in Figure~\ref{fig:base}. Though this is the widely adopted approach, it discards data that are outside the prediction window but are collected in practice as part of the training set. Ideally, these data can be utilized to enhance early prediction, such as learning the future trajectory of the time-series to assist modeling. Leveraging data available at training time but not at test time -- referred to as \textit{privileged information} -- for training has been proposed  as Learning using Privileged Information (LuPI)~\citep{Vapnik2009-xz,Vapnik2015-qt}. Recent studies have shown LuPI can be successfully applied to utilize time-series privileged information for early prediction~\citep{Hayashi2019-cr,KA_Karlsson2022-dc}. However, these experiments focus on structured features from synthetic data or distributions under certain assumptions. It remains unknown whether the approach applies to text-based early prediction applications, where natural language presents distinct characteristics and variation.
 
In this study, we adapt the time-series LuPI to textual data, presented as \textbf{L}earning \textbf{u}sing \textbf{P}rivileged t\textbf{I}me-s\textbf{E}ries \textbf{T}ext (LuPIET). We evaluate LuPIET on a range of tasks to evaluate its efficacy. LuPIET trains a more performant model using a longer predictive window that includes data created after the target prediction point as a teacher model. This is applied to guide the training of the early model 
through knowledge distillation. Figure~\ref{fig:lupiet} gives an example where the prediction window is \oday, and we aim to guide the training of the early model with the teacher model trained from a  \tday window. To compare with LuPIET, we also apply two other methods, common in other domains but not well-examined for text-based time-series, to leverage data collected after the prediction time but available for training. These are transfer learning and mixed training,  depicted in Figure~\ref{fig:transfer} and Figure~\ref{fig:mixed}, respectively.
 
We examine LuPIET using two challenging and domain-specific datasets containing clinical and social media text. Specifically, we explore four risk and diagnosis prediction tasks with clinical text and two mental health status prediction tasks with social media. The results show LuPIET can be an effective and stable approach for improving early prediction based on textual input.   
 
In summary, this work examines the usefulness of privileged time-series information in the NLP context to support early prediction. Our main contributions include:

\begin{enumerate}
    \item Proposing LuPIET to improve time-series modeling for text-based early prediction.
    \item Evaluating the performance of LuPIET on two domains using clinical and social media corpora, presenting results on six prediction tasks. We show that when the privileged text is appropriately chosen and represented, LuPIET can improve over the baseline for early prediction by being more sample efficient.
    \item Benchmarking the performances of two other competitive methods to support early prediction. We find although they can sometimes outperform LuPIET, they are in certain cases detrimental to modeling.  LuPIET offers more consistent and stable improvements over the baseline.
\end{enumerate}

%% file: sections/related_work.tex
\textbf{Early prediction with text} Forecasting or early prediction has been widely studied in various domains and applications. For example, \citet{Steyerberg2009-jw} demonstrates the different facets and modeling strategies for clinical prediction modeling. Most initial works in the field focus on structured measurements as input features, with some attempts to extract and include shallow textual features or topics~\citep{Suresh2017-ql,Ghassemi2015-iy,Ghassemi2014-wq}. More recent studies aim to put more stress on text by applying more powerful models to handle the complexity of language~\citep{Matero2020-tv,Seinen2022-vr}. They have shown promise in modeling various types of text to support the prediction of mental health issues~\citep{Halder2017-qc}, stock market trends~\citep{Xu2018-zy}, and clinical outcomes~\citep{Hsu2020-ga}.

\textbf{Learning using privileged information} LuPI presents a framework to leverage features only available at the train time but not at test time~\citep{Vapnik2009-xz}. It has shown improved results in a range of applications, including recommendation~\citep{Xu2020-br} and image processing~\citep{Lee2020-ie}. Recently, the approach has been applied to improve early prediction using time-series data, which leverages data observed between the prediction time and the future outcome as privileged information. \citet{Hayashi2019-cr} examines this approach on a synthetic dataset and a real-world dataset on air conditions with eight variables. \citet{KA_Karlsson2022-dc} further formalizes the framework for time-series as Learning using Privileged Time-Series (LuPTS) and proves it is guaranteed to result in more efficient learning when the time-series are drawn from a non-stationary Gaussian-linear dynamic system. However, none of the prior works examines text as input. 
 
\textbf{Knowledge distillation} In knowledge distillation (KD), a more performant teacher model guides a smaller student model to achieve better results by matching the distributions of their predictions or output logits~\citep{Hinton2015-sh}. By training with the teacher output, the student model is provided with soft targets that contain more nuanced information about the label distribution compared to the true, hard labels. KD has been widely used for model compression~\citep{Sanh2019-yq,Tung2019-nu,Jiao2020-vs} and other machine learning applications~\citep{Furlanello2018-fb,Clark2019-ji} to transfer knowledge across models with different strengths, sizes, or even architectures. In contrast, the classic transfer learning focuses on a single model and transfers knowledge across datasets, often from larger datasets to smaller ones~\citep{Devlin2019-zp}. Early works have shown the connection between LuPI and KD, unified them under \textit{generalized distillation}~\citep{Lopez-Paz2016-da}. Distillation has become a standard implementating technique to leverage privileged information~\citep{Hayashi2019-cr,Xu2020-br}.

%% file: sections/method.tex
Here we present the problem setting for early prediction and then describe how learning using privileged time-series text (LuPIET) works. 

\subsection{Problem setting}

The goal for early prediction is to learn a mapping function $f(\theta)$ between input $X_{t,n} \in \mathcal{R}^{n\times d}$ and future events or outcomes $Y \in \mathcal{R}$, where $t=1\ldots T$ is a time point of the time-series defined by the prediction window,  and $n=1\ldots N$ is a textual note or post available in the window $X_t$. Since $N$ can vary across prediction windows, we neglect the notation of $n$ from now on for simplicity. Note that $[X_1, X_2,...,X_T]$ share the same label $Y$ as they come from the same sample, where $X_{t}$ is always a subset of $X_{t+1}$. We assume the \textit{baseline} prediction window by setting $t=1$, and the baseline model trained in Figure~\ref{fig:base} is obtained as 

\begin{equation}\label{eq:base}
    \theta_{base} = \mathop{\arg \min}\limits_{\theta \in \Theta} \mathcal{H}(f(X_1), Y)
\end{equation}

where $\mathcal{H}$ is the cross entropy loss. We then aim to improve $\theta_{base}$ by leveraging texts created chronologically after $X_1$, namely $[X_2, X_3, ... X_T]$. 

\subsection{Learning with privileged time-series text}

LuPIET optimizes a knowledge distillation loss that maps the predictions between the baseline model and the new model trained with privileged text, which can be viewed as a teacher model. We train this teacher model using input from a prolonged prediction window compared to baseline, namely $X_t$ where $t\ge2$, to obtain 

\begin{equation}\label{eq:late}
    \theta_{t} = \mathop{\arg \min}\limits_{\theta \in \Theta} \mathcal{H}(f(X_t), Y)
\end{equation}

Then let $p_{base}(x)$ and $p_t(x)$ be the output logits from the base model and teacher model, and we scale them with a temperature $\tau$ before taking the softmax, as defined in the original setting of knowledge distillation~\citep{Hinton2015-sh}:

\begin{align}
    p_{base}^{\tau}(x_i) &= \frac{e^{p_{base}(x_i) / \tau}}{\sum_{j=1}^K e^{p_{base}(x_j) / \tau}} \\
    p_t^{\tau}(x_i) &= \frac{e^{p_t(x_i) / \tau}}{\sum_{j=1}^K e^{p_t(x_j) / \tau}} 
\end{align}

where $K$ is the number of labels. We calculate the distillation loss as the KL-divergence between the two scaled logit distributions as 

\begin{equation}\label{eq:kd}
    \mathcal{L}_{KD} = \mathcal{D}_{KL}(p_{base}^{\tau}(x)||p_t^{\tau}(x))
\end{equation}

This distillation loss is then added to the cross entropy loss to train the final model that consumes the baseline input $X_1$, which is obtained by optimizing the following training objective: 
\begin{equation}
    \mathcal{L} =  (1-\alpha)\mathcal{H}(f(X_1), Y) + \alpha \mathcal{L}_{KD}
\end{equation}
where $\alpha$ is a hyperparameter and $0 \le \alpha \le 1$. 

\subsection{Other options to improve early baseline}
Besides LuPIET, we also examine two other simple methods to leverage privileged text to assist early training --- transfer learning and mixed training. Transfer learning~\citep{Zhuang2021-nk} refers to further fine-tuning $\theta_{t}$ using $X_1$ by minimizing the standard cross entropy loss. In other words, we initialize the training in Eq~\ref{eq:base} using the model parameters from Eq~\ref{eq:late}, as shown in Figure~\ref{fig:transfer}. 

In the mixed training method, the approach is to mix $X_t$ with $X_1$ and train the model from scratch.  This can be considered as a data augmentation approach~\citep{Wen2021-sn} which enriches the training set and  encourages the model to learn from all variations of the same sample. This is depicted in Figure~\ref{fig:mixed}.

%% file: sections/experiment.tex
We examine LuPIET using datasets from two challenging textual domains: clinical notes and social media posts. All datasets contain notes or posts that are created chronologically, naturally forming the text-based time series. Here we introduce them in more detail.

\subsection{Clinical datasets and tasks}
For clinical text, we use the MIMIC-III database~\citep{Johnson2016-kl} to construct the datasets to predict four clinical outcomes and targets, which are in-hospital mortality, in-ICU mortality, length-of-stay (LOS) over 3 days, and diagnostic related groups (DRG). These are popular tasks in the literature for clinical early prediction and we follow previous works to define and extract cohorts for them~\citep{Wang2020-py, Liu2021-jx,Liu2022-rp}. We note that the early DRG prediction is different from the typical medical coding performed post-discharge~\citep{Dong2022-sb,Liu2022-jb} as it aims to predict diagnosis and estimate care costs while patients are still in the hospital~\citep{Gartner2015-lh,Islam2021-tg}. 

For each patient in the cohort, we extract all clinical notes during the hospital course that are created before the prediction time, sort them chronologically, and remove empty or duplicated notes. The prediction window is defined by the number of days after the patient ICU admission. For example, the \tday window would include all clinical notes charted by the end of the third day of ICU admission.

We define the baseline window for the clinical prediction tasks as \oday, which would allow timely interventions and resource arrangements to be made for the hospitalized patients. This is also a common choice made in the literature~\citep{Wang2020-py,Hsu2020-ga}. We then examine two extended prediction windows to train LuPIET, which are \tday and \sday. Notice there are cases whose time-series lengths are less than \tday or \sday. We did not distinguish these cases from others in the data extraction process to ensure we can directly compare with baseline results. For example, if a case has a length of 2 days, then the input texts for this case under \tday and \sday are the same. The numbers of train/validation/test cases are presented in Table~\ref{tab:stat}. Notice the two mortality predictions and LOS prediction share the same cohort.  

\subsection{Social media datasets and tasks}
For social media texts, we focus on the eRisk 2018 datasets~\citep{Losada2016-hb,Losada2018-lh} to use Reddit posts to predict potential mental health issues, which are depression and anorexia. Predicting mental health status using social media data is an important yet challenging task~\citep{Guntuku2017-tw} that draws much attention in recent years from the NLP community~\citep{Benton2017-ri,Cohan2018-ca}. To parse the datasets, we use both the title and the content as input text for the post. The datasets present the posts in ten chunks, which are evenly split by time and sorted in the chronological order. We follow this format to define the prediction window by the number of chunks used as input, e.g., a \tchunk window includes all posts in the first three chunks.

Since social media text could present noisier temporal patterns, we examine two baseline windows for the prediction tasks, which are \tchunk and \schunk, and only use the full \tenchunk as prolonged prediction window. We make this choice also because the eRisk datasets are relatively small and the performance can be unstable. We use all samples from the datasets for each task and split them in a 0.8/0.1/0.1 ratio, and the numbers are again presented in Table~\ref{tab:stat}.

\begin{table}[]
    \small
    \resizebox{\linewidth}{!}{
    \begin{tabular}{lcccc}
    \hline
       & \# train  & \# validation  & \# test  & \# labels \\ \hline
        
      Mortality \& LOS   & 26729 & 3407 & 3392 & 2 \\ 
      DRG & 16296 & 972 & 1866 & 570 \\ \hline
      eRisk Depression & 656 & 82 & 82 & 2 \\ 
      eRisk Anorexia & 376 & 47 & 48 & 2 \\ \hline
    \end{tabular}
    }
    \caption{Number of cases for the train, validation, and test sets of the examined prediction tasks.}
    \label{tab:stat}
\end{table}

\subsection{Text representation and modeling}
We examine two text representations for the text-based time-series modeling. The first one is to model all input text as a single text string by concatenating all notes or posts together. This allows us to model the input as a sequence of words, which considers the word-level details. This is a standard practice in NLP. The other representation is to encode all notes or posts into document embeddings and model them at the document level. This method, on the other hand, may lose details during the document encoding process, but it helps to maintain the temporal 
patterns of the texts. 

For word-level representation, we use domain-specific pretrained word embeddings for the two datasets. Specifically, we use BioWordVec~\citep{Zhang2019-ua} for clinical text and GloVe-840B~\citep{Pennington2014-pi} for social media. Since the concatenated text string is rather long, we adapt MultiResCNN~\citep{Li2020-dm} for modeling, which is a CNN-based model that has shown strong results in long-document classification. The model enhances the vanilla text-CNN model by adding more filters with residual connections. Given space restrictions here, we do not introduce the architecture in detail and refer readers to the original paper for more information.

For document-level modeling, we first encode notes or posts using BERT~\citep{Devlin2019-zp} by extracting the [CLS] token as the final representation. We again try to align BERT with the domains of the text by using ClinicalBERT~\citep{Alsentzer2019-rm} for clinical notes and RoBERTa~\citep{Liu2019-fa} for Reddit posts. We then apply LSTM~\citep{Hochreiter1997-bh} to model the sequence of notes and use the last hidden state for final prediction. We use LSTM due to its power to extract and model temporal patterns.

\subsection{Training and evaluation}

We tune the hyperparameters for all examined methods based on the validation set and then retrain the model with the best configuration with multiple random seeds. Specifically, we tune the filter number and sizes for the CNN model, hidden size for the LSTM model, and the number of layer, dropout, learning rate, weight decay for both models. Adam~\citep{Kingma2014-kl} is used as the optimizer throughout the experiments. After obtaining the best architectural configuration for the baseline model, we fix the setup and tune $\tau$ and $\alpha$ for LuPIET. We adopt random search for the clinical datasets and grid search for social media datasets given the former takes much longer to run. We facilitate the hyperparameter searching with asynchronous successive halving algorithm~\citep{Li2020-np}, based on the implementation from Ray Tune~\citep{Liaw2018-uj}. We perform all our experiments using pytorch~\citep{Paszke2019-yo} and pytorch-lightning\footnote{https://www.pytorchlightning.ai/} on V100 GPUs.

For evaluation, we use area under the receiver operating characteristics curve (AUROC) and precision-recall curve (AUPR) for  the binary classification tasks, and accuracy and macro F1 for multiclass classification. We run the final, tuned model for each task with five random seeds for the clinical datasets and average the results, and run with ten seeds for the social media datasets given the small data size.

%% file: sections/results.tex
\begin{table*}[!h]
\centering

\resizebox{\linewidth}{!}{%
\begin{tabular}{lc cc cc cc c}
\toprule

\multirow{2}{*}{}    & \multicolumn{2}{c}{In-hospital Mortality} & \multicolumn{2}{c}{In-ICU Mortality} & \multicolumn{2}{c}{LOS$>$3days} & \multicolumn{2}{c}{DRG} \\ \cline{2-9}
& AUROC  & AUPR   & AUROC   & AUPR & AUROC   & AUPR &  Acc   & MacroF1   \\ \midrule


\multicolumn{5}{l}{\textbf{\textit{Baseline}}} \\ 
\oday
 & \makecell{0.867 \\ (0.0032)} & \makecell{0.474 \\ (0.0151)} & \makecell{0.862 \\ (0.0110)} & \makecell{0.373 \\ (0.0252)} & \makecell{0.690 \\ (0.0034)} & \makecell{0.625 \\ (0.0047)} & \makecell{0.282 \\ (0.0058)} & \makecell{0.105 \\ (0.0106)}\\

\hline

\multicolumn{5}{l}{\textbf{\textit{LuPIET}}} \\ 

\tday $\rightarrow$ \oday
 & \makecell{0.876 \\ (0.0027)} & \makecell{0.491 \\ (0.0039)} & \textbf{\makecell{0.880 \\ (0.0020)}} & \textbf{\makecell{0.429 \\ (0.0092)}} & \makecell{0.693 \\ (0.0042)} & \makecell{0.626 \\ (0.0058)} & \makecell{0.287 \\ (0.0068)} & \makecell{0.116 \\ (0.0067)}\\

\sday $\rightarrow$ \oday
 & \textbf{\makecell{0.879 \\ (0.0029)}} & \textbf{\makecell{0.501 \\ (0.0093)}} & \makecell{0.880 \\ (0.0024)} & \makecell{0.413 \\ (0.0076)} & \makecell{0.692 \\ (0.0035)} & \textbf{\makecell{0.627 \\ (0.0041)}} & \makecell{0.290 \\ (0.0097)} & \makecell{0.115 \\ (0.0130)}\\

\hline

\multicolumn{5}{l}{\textbf{\textit{Transfer}}} \\ 

\tday $\rightarrow$ \oday
 & \makecell{0.863 \\ (0.0026)} & \makecell{0.466 \\ (0.0063)} & \makecell{0.866 \\ (0.0060)} & \makecell{0.381 \\ (0.0231)} & \textbf{\makecell{0.695 \\ (0.0036)}} & \makecell{0.612 \\ (0.0049)} & \makecell{0.293 \\ (0.0033)} & \makecell{0.134 \\ (0.0059)}\\

\sday $\rightarrow$ \oday
 & \makecell{0.865 \\ (0.0040)} & \makecell{0.482 \\ (0.0073)} & \makecell{0.860 \\ (0.0051)} & \makecell{0.379 \\ (0.0089)} & \makecell{0.691 \\ (0.0046)} & \makecell{0.616 \\ (0.0068)} & \makecell{0.284 \\ (0.0027)} & \makecell{0.113 \\ (0.0072)}\\
 
\sday $\rightarrow$ \tday $\rightarrow$ \oday
 & \makecell{0.864 \\ (0.0040)} & \makecell{0.482 \\ (0.0110)} & \makecell{0.855 \\ (0.0073)} & \makecell{0.374 \\ (0.0098)} & \makecell{0.683 \\ (0.0031)} & \makecell{0.606 \\ (0.0054)} & \makecell{0.285 \\ (0.0093)} & \makecell{0.112 \\ (0.0081)}\\

\hline
\multicolumn{5}{l}{\textbf{\textit{Mix-train}}} \\ 
\oday + \tday + \sday

 & \makecell{0.866 \\ (0.0031)} & \makecell{0.490 \\ (0.0031)} & \makecell{0.860 \\ (0.0079)} & \makecell{0.398 \\ (0.0077)} & \makecell{0.642 \\ (0.0037)} & \makecell{0.561 \\ (0.0057)} & \textbf{\makecell{0.298 \\ (0.0025) }} & \textbf{\makecell{0.140 \\ (0.0081)}} \\

\bottomrule


\end{tabular}}
    \caption{Results of word-level modeling for the four clinical prediction tasks.}
    \label{tab:res_main_word}
\end{table*}

\begin{table}[]
    \small
    \resizebox{\linewidth}{!}{%
    \begin{tabular}{l cc cc }
    \toprule
    \multirow{2}{*}{}    & \multicolumn{2}{c}{Depression} & \multicolumn{2}{c}{Anorexia}  \\ \cline{2-5} 
& AUROC  & AUPR   & AUROC  & AUPR   \\ \midrule
     Baseline - \tchunk    & \makecell{0.819 \\ (0.0636)} & \textbf{\makecell{0.458 \\ (0.1606)}}  & \makecell{0.796 \\ (0.0742)} & \makecell{0.334 \\ (0.1071)}  \\
      \makecell{  + \textbf{\textit{LuPIET}}}  & \textbf{\makecell{0.868 \\ (0.0203)}} & \makecell{0.431 \\ (0.1150)} & \textbf{\makecell{0.830 \\ (0.0742)}} & \textbf{\makecell{0.339 \\ (0.0963)}}  \\
      
      \hline
      
      Baseline - \schunk & \makecell{0.836 \\ (0.0303)} & \makecell{0.470 \\ (0.1374)} & \makecell{0.798 \\ (0.0401)}	& \textbf{\makecell{0.429 \\ (0.1217)}} \\
      \makecell{  + \textbf{\textit{LuPIET}}}  & \textbf{\makecell{0.869 \\ (0.0332)}} & \textbf{\makecell{0.495 \\ (0.0792)}}  & \textbf{\makecell{0.807 \\ (0.0239)}} & \makecell{0.423 \\ (0.0705)}  \\
      
     \bottomrule
    \end{tabular}}
    \caption{Results for the two mental health prediction tasks using social media posts.}
    \label{tab:res_social}
\end{table}

\subsection{Comparing LuPIET with baseline}\label{sec:res}

We present our main results with the word-level modeling for clinical prediction tasks in Table~\ref{tab:res_main_word}. We observe LuPIET using the extended \tday and \sday windows improves over the standard baseline window on \oday for in-hospital mortality, in-ICU mortality, and DRG predictions. Meanwhile, LuPIET does not provide much benefit for predicting LOS$>$3days. We believe this is related to the nature of the task as the extended windows are already longer than the target in consideration, so the teacher models can take advantage of this shortcut to make accurate and confident predictions that are close to the true labels and can no longer serve as soft targets~\citep{Hinton2015-sh,Cho2019-du}. 

The results for social media datasets are presented in Table~\ref{tab:res_social}. Here we examine \tchunk and \schunk as the baseline windows and apply LuPIET trained using full length, i.e., \tenchunk. We again observe the benefit of LuPIET over the baseline results, especially under AUROC. However, we note that due to the much smaller dataset size, the variances in the results can be large. In a couple of cases, LuPIET achieves slightly lower AUPR scores than the baseline, but the results are still comparable given the variance.

\subsection{Comparing LuPIET with other methods} 
We focus on the clinical datasets to compare LuPIET with transfer learning and mixed training approaches given they have relatively sufficient data to observe their behavior. In Table~\ref{tab:res_main_word}, we find LuPIET still performs strongly when compared to these two approaches, but other methods may outperform LuPIET, such as mixed training on DRG prediction. In this particular case, we believe this is caused by the DRG dataset being a multiclassification task with 570 labels, thus not having enough samples for training. By mixing different windows of the same sample together serves as a data augmentation strategy, which alleviates the low-resource situation for DRG to achieve better results. Similarly, when modeling at the note level (Table~\ref{tab:res_note_rep}), transfer learning may be able to better transfer the temporal relations across windows thus achieving slightly better results. 

However, we observe that transfer learning and mixed training do not always improve over the baseline, which is further shown when we examine the note-level modeling results in Table~\ref{tab:res_note_rep}. For example, though it could bring benefits to tasks like DRG, mixed training is rather detrimental to the LOS$>$3days prediction. On the other hand, LuPIET either improves over the baseline or at least maintains the performance under both text representation and modeling strategies. 

We also see LOS$>$3days task does not benefit much from transfer learning and mixed training either, similar to the results with LuPIET. This shows when adopting these methods to improve the time-series modeling, it can be important to consider the nature of the task and to choose proper windows accordingly.

\subsection{Comparing text representations} 
The results presented in Table~\ref{tab:res_main_word} and Table~\ref{tab:res_note_rep} for word- and note-level modeling are comparable for all the four tasks. Overall, we see modeling at word level performs much better than at note level, demonstrating the need to attend to the fine-grained textual details for these tasks. The LOS$>$3days is again an exception where the two sets of results are similar, showing the task can be inherently more difficult. Furthermore, modeling at word level tends to benefit more from LuPIET. For example, we see much better mortality prediction scores with LuPIET in Table~\ref{tab:res_main_word}. This indicates that the proper training of LuPIET exploits the nuances in the data and it would benefit more when modeling at a finer granularity. 

We do not present the results with document embeddings for the social media tasks as their AUROC results are sub-optimal and barely over 0.5. We suspect this is because the Reddit posts are much noisier and the model needs to sift away much unrelated information, so encoding all posts into embeddings is unhelpful.

\begin{table*}[h!]
\centering
\small

\resizebox{\textwidth}{!}{%
\begin{tabular}{lc cc cc cc c}

\toprule

\multirow{2}{*}{}    & \multicolumn{2}{c}{In-hospital Mortality} & \multicolumn{2}{c}{In-ICU Mortality} & \multicolumn{2}{c}{LOS$>$3days} & \multicolumn{2}{c}{DRG} \\ \cline{2-9} 
& AUROC  & AUPR   & AUROC   & AUPR & AUROC   & AUPR &  Acc   & MacroF1   \\ \midrule

\multicolumn{5}{l}{\textbf{\textit{Baseline}}} \\ 
\oday 

& \makecell{0.844 \\ (0.0058)} & \makecell{0.417 \\ (0.0122)} & \makecell{0.860 \\ (0.0049)} & \makecell{0.369 \\ (0.0172)} & \makecell{0.695 \\ (0.0059)} & \makecell{0.629 \\ (0.0061)} & \makecell{0.228 \\ (0.0057)} & \makecell{0.056 \\ (0.0043)}\\

\hline

\multicolumn{5}{l}{\textbf{\textit{LuPIET}}} \\ 

\tday $\rightarrow$ \oday
 & \makecell{0.850 \\ (0.0040)} & \makecell{0.428 \\ (0.0108)} & \makecell{0.861 \\ (0.0060)} & \makecell{0.362 \\ (0.0178)} & \makecell{0.696 \\ (0.0034)} & \makecell{0.631 \\ (0.0054)} & \textbf{\makecell{0.238 \\ (0.0063)}} & \makecell{0.056 \\ (0.0055)}\\

\sday $\rightarrow$ \oday
 & \makecell{0.851 \\ (0.0033)} & \makecell{0.430 \\ (0.0125)} & \makecell{0.860 \\ (0.0046)} & \makecell{0.370 \\ (0.0168)} & \textbf{\makecell{0.698 \\ (0.0027)}} & \textbf{\makecell{0.635 \\ (0.0039)}} & \makecell{0.237 \\ (0.0042)} & \makecell{0.057 \\ (0.0029)}\\
 
\hline

\multicolumn{5}{l}{\textbf{\textit{Transfer}}} \\ 

\tday $\rightarrow$ \oday
 & \textbf{\makecell{0.853 \\ (0.0038)}} & \textbf{\makecell{0.439 \\ (0.0080)}} & \makecell{0.862 \\ (0.0069)} & \textbf{\makecell{0.384 \\ (0.0093)}} & \makecell{0.688 \\ (0.0022)} & \makecell{0.615 \\ (0.0044)} & \makecell{0.232 \\ (0.0067)} & \makecell{0.059 \\ (0.0043)}\\

\sday $\rightarrow$ \oday
 & \makecell{0.833 \\ (0.0037)} & \makecell{0.415 \\ (0.0086)} & \makecell{0.861 \\ (0.0065)} & \makecell{0.371 \\ (0.0123)} & \makecell{0.686 \\ (0.0056)} & \makecell{0.623 \\ (0.0057)} & \makecell{0.230 \\ (0.0070)} & \makecell{0.064 \\ (0.0039)}\\

\sday $\rightarrow$ \tday $\rightarrow$ \oday
 & \makecell{0.829 \\ (0.0073)} & \makecell{0.405 \\ (0.0108)} & \makecell{0.861 \\ (0.0080)} & \makecell{0.362 \\ (0.0147)} & \makecell{0.688 \\ (0.0029)} & \makecell{0.621 \\ (0.0022)} & \makecell{0.232 \\ (0.0052)} & \textbf{\makecell{0.066 \\ (0.0049)}}\\

\hline

\multicolumn{5}{l}{\textbf{\textit{Mix-train}}} \\ 
\oday + \tday + \sday

 & \makecell{0.838 \\ (0.0019)} & \makecell{0.416 \\ (0.0113)} & \textbf{\makecell{0.866 \\ (0.0055)}} & \makecell{0.382 \\ (0.0125)} & \makecell{0.671 \\ (0.0035)} & \makecell{0.605 \\ (0.0066)} & \makecell{0.236 \\ (0.0073)} & \makecell{0.061 \\ (0.0043)}\\

\bottomrule

\end{tabular}}
    \caption{Results of document-level modeling for the four clinical prediction tasks.}
    \label{tab:res_note_rep}
\end{table*}

%% file: sections/discussion.tex

\begin{figure*}[!h]
    \centering
    \includegraphics[width=.9\textwidth]{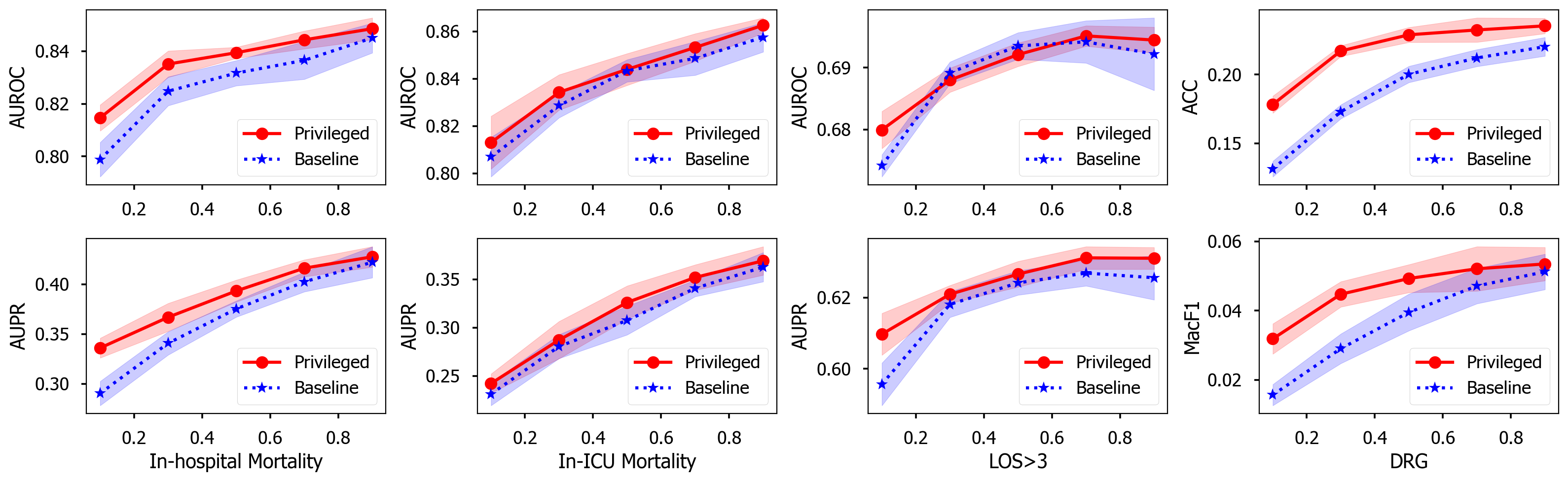}
    \caption{Learning curves for the four clinical prediction tasks under two evaluation metrics. The blue curves depict the results of standard training and the red curves depict those of LuPIET.}
    \label{fig:curve}
\end{figure*}

\subsection{Sampling efficiency}
\citet{KA_Karlsson2022-dc} shows under certain conditions, such as when the time-series is from linear dynamical systems with Gaussian noise and is in Markov structure, privileged information is guaranteed to improve the learning efficiency of time-series models. Given these conditions do not necessarily hold for the natural language that has distinctive data distribution, here we empirically examine if privileged text can still benefit NLP modeling efficiency without making other assumptions. Figure~\ref{fig:curve} shows the learning curves for the four clinical prediction tasks, with x-axis is the ratio of the full dataset used to train the model. We find LuPIET can lead to more sample efficient learning in multiple tasks. For example, when training with only 10\% of the whole dataset, privileged learning achieves significant improvements over the baseline on in-hospital mortality, LOS, and DRG. When the model is fed with more samples, we see the difference between LuPIET and the baseline gradually converge. This also happens to the much smaller social media datasets, where we see the larger extent of improvements with LuPIET under AUROC in Table~\ref{tab:res_social}. Furthermore, LuPIET could reduce the modeling variance compared to baseline, consistent with findings on structured time-series~\citep{KA_Karlsson2022-dc}.

Meanwhile, in certain scenarios, such as for the in-ICU mortality prediction and for the LOS prediction with sufficient data, we find the benefits of LuPIET are much weaker. This  may reflect our observations on LOS in Sec~\ref{sec:res}, where the choice of input windows can be important. Extending input windows in certain scenarios may not necessarily include more data. For instance, in the ICU admissions stays are much shorter compared to hospital stays. This may explain why little benefit is observed for in-ICU mortality prediction.

\subsection{Limitations}
There are a few limitations with our study. Firstly, we find it is important to apply LuPIET to appropriate task settings but we did not formalize the definition of appropriateness, though we offer some possible intuitions (e.g., on the case of LOS in Sec~\ref{sec:res}). Domain knowledge about the nature of the task may be needed to realize optimal results with LuPIET, which could in some ways correspond to the assumptions about data distribution made for successful LuPI in previous works~\citep{Lopez-Paz2016-da,KA_Karlsson2022-dc}. Future works are needed to provide the theoretical explanation for the empirical results and to guide more effective application of LuPIET.

Secondly, we find sometimes the teacher models do not benefit from extending the input window and consuming more time-series data, for example, the shorter \tchunk can outperform \schunk for anorexia prediction (Table~\ref{tab:res_social}). We did not further investigate this phenomenon and leave it to future work to explore the potential causes. We also focused on the specific time steps for baseline and extended input windows and did not evaluate in more time steps, but in the future we would like to consider various time steps in the time-series for both training and evaluation. Similar evaluation setup has been examined in prior work~\citep{Harutyunyan2019-qr}, such as framing patient deterioration assessment as hourly patient mortality predictions. We regard this setup a promising extension of our current experiments to examine LuPIET. Lastly, we do not explore how factors in successful KD applications~\citep{Gou2021-tw}, such as creating consistent teacher models~\citep{Beyer2022-cv}, affect LuPIET, which we consider as a future direction to better utilize privileged information and to enhance LuPI in general.

%% file: sections/conclusion.tex
In this study, we present LuPIET, a framework to 
incorporate longer-range 
time-series data
available during training 
to improve text-based early predictions. Though similar ideas have been examined recently for  structured time-series~\citep{Hayashi2019-cr,KA_Karlsson2022-dc}, we are not aware of any previous studies 
on the use of this 
privileged information in the context of text-based time-series. We find LuPIET is an effective strategy for enhancing early prediction and for efficient time-series modeling when applied to appropriate task settings. 

LuPIET is implemented by simply optimizing a distillation loss. Therefore, future works may extend LuPIET by training with more advanced distillation techniques, e.g., matching hidden state instead of logits~\citep{Zhang2018-gg}, or combining with other inputs, e.g., using multi-modal privileged information.